\documentclass[lettersize,journal]{IEEEtran}
\usepackage{amsmath,amsfonts}
\usepackage{algorithmic}
\usepackage{algorithm}
\usepackage{array}
\usepackage{float}
\usepackage[
 font=footnotesize
 ]{subfig}
\usepackage{textcomp}
\usepackage{stfloats}
\usepackage{url}
\usepackage{verbatim}
\usepackage{graphicx}
\usepackage{cite}
\usepackage{graphicx}
\usepackage{booktabs}
\usepackage{multirow}
\usepackage[utf8]{inputenc}
\usepackage{pifont}
\usepackage{newunicodechar}
\usepackage[pagebackref,breaklinks,colorlinks]{hyperref}

\usepackage{orcidlink}

\usepackage{ifthen} 
\newboolean{review} 
\setboolean{review}{false} 

\ifthenelse{\boolean{review}}{
\usepackage{ulem}
\newcommand{\fix}[1]{{\color{blue}#1}} % Add a new blue text
\newcommand{\fixS}[1]{{\color{blue}\sout{#1}}} % Scratch old replace with new text
\newcommand{\fixM}[2]{\fix{#1\marginpar{\color{red}#2}}} % Add new blue text and a pointer to it on the side
}{
\newcommand{\fix}[1]{#1}
\newcommand{\fixS}[1]{}
\newcommand{\fixM}[2]{#1}
}
\begin{document}

\title{HyperspectralViTs: General Hyperspectral Models for On-board Remote Sensing}

\author{Vít Růžička \orcidlink{0000-0001-6558-7197}, Andrew Markham \orcidlink{0000-0001-5716-3941} % <-this % stops a space
\thanks{University of Oxford, Oxford, UK. Email: firstname.lastname@cs.ox.ac.uk \textit{(Corresponding author: Vít Růžička.)}}% <-this % stops a space
\thanks{Manuscript received October 24, 2024; revised January 9, 2025.}
\thanks{Accepted manuscript pre-print. Published version may differ.}
\thanks{Code is available at: \url{https://github.com/previtus/HyperspectralViTs}}
}

% \markboth{IEEE JOURNAL OF SELECTED TOPICS IN APPLIED EARTH OBSERVATIONS AND REMOTE SENSING,~Vol.~??,~2025}%
\markboth{IEEE JOURNAL OF SELECTED TOPICS IN APPLIED EARTH OBSERVATIONS AND REMOTE SENSING,~2025}%
% \markboth{Preprint. Under review.}%
{Ruzicka \MakeLowercase{\textit{et al.}}: General Hyperspectral Models for On-board Remote Sensing}
% IEEE TRANSACTIONS ON GEOSCIENCE AND REMOTE SENSING,~Vol.~??,~2024
% IEEE JOURNAL OF SELECTED TOPICS IN APPLIED EARTH OBSERVATIONS AND REMOTE SENSING,~Vol.~??,~2024

% \IEEEpubid{0000}
% \IEEEpubid{0000--0000/00\$00.00~\copyright~2025 IEEE}
\IEEEpubid{\copyright~2025 The Authors. This work is licensed under a Creative Commons Attribution 4.0 License. For more information, see \url{https://creativecommons.org/licenses/by/4.0/}}

\maketitle

% Limit to 250 words:

\begin{abstract}
    \fixM{On-board processing of hyperspectral data with machine learning models would enable an unprecedented amount of autonomy across a wide range of tasks allowing new capabilities such as early warning systems and automated scheduling across constellations of satellites. 
    However, current classical methods suffer from high false positive rates and therefore prevent easy automation while previously published deep learning models exhibit prohibitive computational requirements.
    We propose fast and accurate machine learning architectures which support end-to-end processing of data with high spectral dimension without relying on hand-crafted products or spectral band compression techniques. 
    We create three new large datasets of hyperspectral data containing all relevant spectral bands from the near global sensor EMIT.
    We evaluate our models on two tasks related to hyperspectral data processing - methane detection and mineral identification.
    Our models reach a new state-of-the-art performance on the task of methane detection, where we improve the F1 score of previous deep learning models by 27\% on a newly created synthetic dataset and by 13\% on the previously released large benchmark dataset.
    Our models generalise from synthetic datasets to data with real methane leak events and boost performance by 6.9\% in F1 score in contrast with training models from scratch on the real data.
    Finally, with our newly proposed architectures, one capture from the EMIT sensor can be processed within 30 seconds on a realistic proxy of the ION-SCV 004 satellite and in less than 0.64 seconds on a GPU powered Jetson AGX Xavier board.
    }{R2Q1}
\end{abstract}


\section{Introduction}
\label{sec:intro}

% // intro – new ml model
\IEEEPARstart{I}{n} this paper, we introduce HyperspectralViTs, new adaptations of Transformer based machine learning models for semantic segmentation of hyperspectral data (also refered to as imaging spectroscopy data) for low compute environments. We demonstrate these adaptations on two recent architectures, SegFormer \cite{xie2021segformer} and EfficientViT \cite{cai2023efficientvit_base} and for two relevant tasks, methane leak detection and mineral identification. 
We describe the existing landscape of approaches used to detect methane leaks and minerals in remote sensing data and outline several limitations to these methods. 
We then use these as motivation to propose changes to the Transformer based architectures. We improve the accuracy and inference speed of the existing methods and benchmark the model in limited compute environment which serves as a proxy to the hardware available on the satellite.

% // motivation: methane detection - important
Methane detection is an important task, which could lead to the reduction of anthropogenic methane in atmosphere - this has been identified as one of the fastest pathways to reduce global warming in United Nations Global Methane Assessment report \cite{kuylenstierna2021global_UN}.
In practice, the detection of methane leaks remains to be manual, as the outputs of existing methods can be quite noisy. For large scale automation of this process, reliable and accurate models are needed to sift through the increasing deluge of data from newly deployed hyperspectral sensors.

Mineral identification is a task related to scene decomposition into individual endmembers from a spectral library. On its own it could serve as a strong general task highly relevant for future Foundation models working with hyperspectral data. 
\fixM{Hyperspectral Foundation models have been explored in the very recent works of \cite{braham2024spectralearth_foundation_hsi, wang2024hypersigma_foundation_hsi}.}{R1Q6}
We note that one of the example downstream tasks of interest could also be methane detection, as it has been shown that having information about scene background is beneficial for separation of methane leak events from confounders \cite{kumar2023methanemapper}.

\IEEEpubidadjcol

% // motivation: on-board ML
Autonomy on-board of satellites is needed to enable rapid response, alerting and automatic scheduling within a constellation of satellites \cite{furano2020towards, parr2024nio_trillium}. Recent works have shown that detection of events of interest (such as floods) can be done directly in space with on-board machine learning models, reducing the quantity of data for downlinking \cite{mateo2021towardsFlooding}.

We note that the proposed architecture is aimed at the general problem of processing hyperspectral data and as such can be used outside of the domain of remote sensing. By framing the problem as an end-to-end semantic segmentation, we disentangle the proposed architecture from the task of methane leak detection and retain generality. 

Developing machine learning architecture for hyperspectral data is well timed and necessary for many downstream applications.
Hyperspectral remote sensing data is used in mineralogy \cite{peyghambari2021hyperspectral_mineralogy} and agriculture \cite{lu2020recent_agriculture}. Sensors with high spectral dimension are deployed on devices used for deep space exploration \cite{blaney2019europa_hyperspectral_mise, ehlmann2021lunar_trailblazer_spectral, green2011moon_earlyformoon} and on hand-held devices \cite{ravikumar2017optical_handheld} including mobile phones \cite{spectracity_news}.

\begin{figure}[h]
    \centering
    \includegraphics[width=1.0\linewidth]{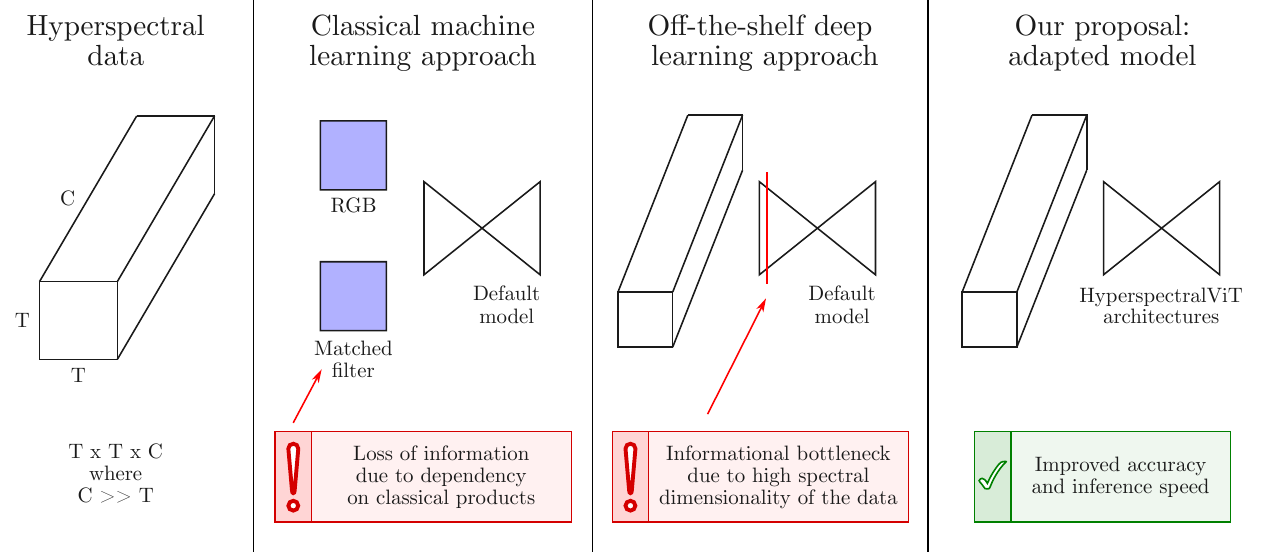}
    \caption{Illustration of the limitations of previous approaches. \fix{First, dependence on classically used products computed from hyperspectral data (such as matched filters in \cite{STARCOP}) may lead to loss of accuracy due to imperfect capture of the events of interest. Second, off-the-shelf versions of machine learning models aren’t adapted to hyperspectral data and may lose valuable information in the early layers of the models. Finally, our adapted model leverages information from all relevant hyperspectral bands, which leads to improvements in both accuracy and inference speeds of our method.}}
    \label{fig:contribution}
\vspace{-2mm}%less space in between figs 
\end{figure}

Our main contribution is the proposal of a new family of models called HyperspectralViTs, architectures adapted to handle hyperspectral data without informational losses due to reliance on classically computed intermediate products, or due to informational bottlenecks in the original architectures as illustrated on Figure \ref{fig:contribution}.
We analyse the limitations of existing machine learning architectures without adaptations for hyperspectral data and show that we can improve their performance with several simple steps. 

Namely, we propose the HyperSegFormer and HyperEfficientViT variants of the default models and evaluate them on a newly created dataset of synthetic methane leak events in the data from the near global hyperspectral sensor EMIT \cite{green2022_EMIT}. We then demonstrate that these gains translate over to a newly created dataset of real methane leak events from the same sensor.
We show that we are able to improve the accuracy and the inference speed of the existing classical and machine learning approaches by removing their dependency on the classical matched filter product.

Finally, we demonstrate that our models are relevant also for other tasks of interest on a newly created dataset for mineral identification.

\subsection{Literature review}
\label{sec:lit_rev}

In this paper we primarily tackle the tasks of methane leak detection, as such most of the relevant literature focuses on this discipline. We however also keep a brief section on methods used for mineral identification.

The task of methane leak detection can be more generally described as semantic segmentation of remote sensing images and has recently seen increased interest from the machine learning community \cite{tiemann2024ML4CH4_overview}.

\begin{figure}[h]
    \centering
    \includegraphics[width=0.9\linewidth]{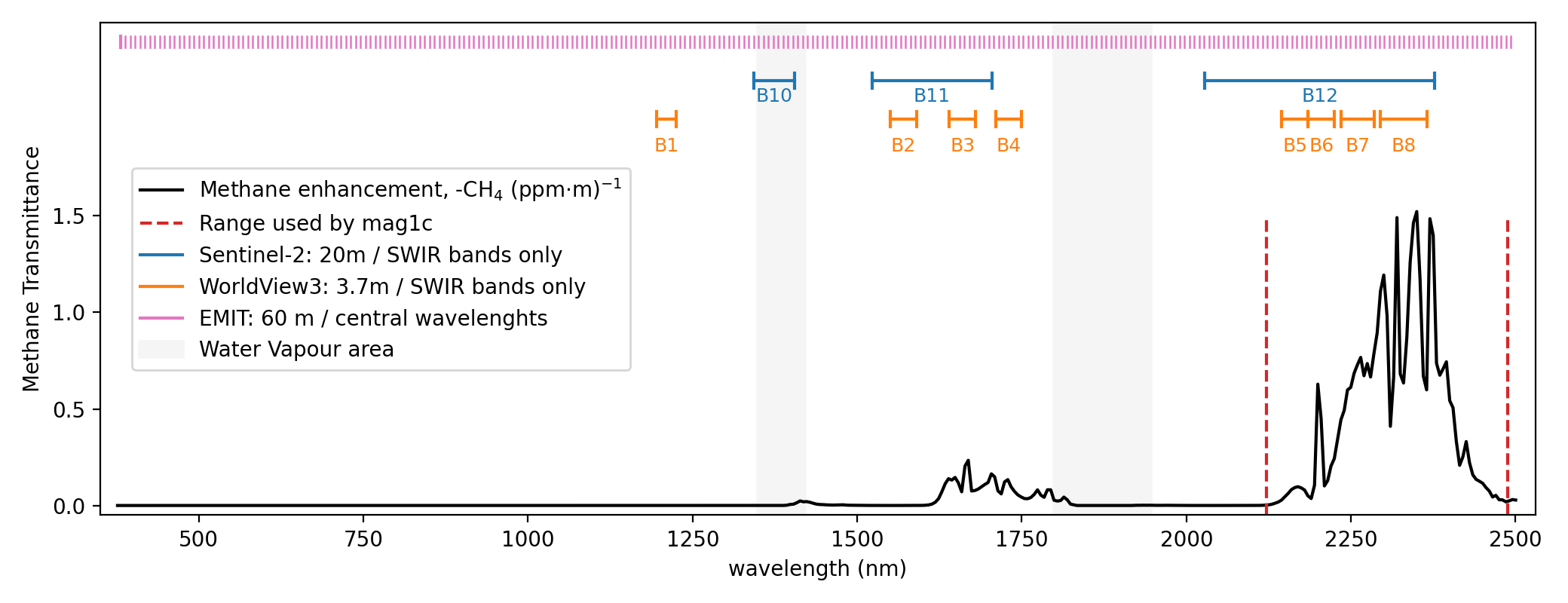}
    \caption{Methane gas signature (shown through the methane transmittance) in comparison with the band ranges of typically used multispectral and hyperspectral satellites.}
    \label{fig:methaneXsatellites}
\vspace{-2mm}%less space in between figs 
\end{figure}

Figure \ref{fig:methaneXsatellites} shows the wavelength range of typically used multispectral and hyperspectral satellites alongside with the target signature of methane (modelled from the HITRAN 2012 database \cite{BROWN2013201_HITRAN}). The multispectral satellite Sentinel-2 has been previously used to monitor large methane leaks in \cite{varon_high-frequency_2021}, however multi-temporal passes over the same location were needed to detect weaker events. The work of \cite{sanchez-garcia_mapping_2022} uses data from the WorldView-3 satellite, which has multiple bands in the desired area of interest, but only large methane leak events can be detected and validation requires human expert intervention.

Hyperspectral data provides more information for the task of detecting methane signal, this is due to the improved sampling in spectral ranges where methane is visible (namely between 1600-1800nm and 2100-2500nm). This task has been classically addressed with matched filter approaches \cite{Thompson2015_realtime} which compare the observed data against the target signature of methane. The resulting matched filter product, sometimes called  ``methane enhancement'', is quite noisy, as other confounder materials (for example rooftops, roads and solar panels) get detected as well. Iterative matched filter approaches, such as the mag1c filter \cite{foote_fast_2020_mag1c} attempt to improve the clarity of the product. 
Other works depend on matched filter variants using wider spectral windows \cite{roger2023wide_MF}.
Up until recently, these methane enhancement products were manually inspected, and trained experts were tasked to create final detections.

Mineral detection in remote sensing data has been approached with a wide range of classical methods which are described in the overview of \cite{shirmard2022review_minerals}, and only a subset of these uses hyperspectral data.
Mineral identification in hyperspectral data can be formalised as spectral unmixing, that is identifying which individual endmembers from a spectral library of chemical and mineral constituents create the observed signature. 
One classical approach is the Tetracoder program \cite{clark2003imaging_Tetracorder} which uses a cascade of manually derived rules to compare observed spectral feature with a reference spectral library of minerals. It isolates discriminative portions of spectra known for each mineral using continuum removal techniques \cite{clark1999_continuumRemoval} and then computes degree of fit metrics for each item in the library. It has been described as a simulation of the process a trained mineralogist would do when looking at the hyperspectral data.
This task can also be formalised as semantic segmentation, using ``multi-hot'' labels (multiple minerals can be present in each pixel). 

We note that datasets for global mineral identification across the entire globe are missing. Works such as \cite{wan2021application} evaluate their methods on just two airborne hyperspectral captures. Works such as \cite{clark2003imaging_Tetracorder} compare their methods against several well studied scenes such as the Cuprite or Alunite Hill scenes - these have been sparsely in-situ sampled to provide verification data.
Note that large scale differences exist when comparing in-situ measurements and space based instruments (where one pixel can cover over e.g. 60x60m area).
Therefore, existing works mostly remain local in their scope, and the used datasets are quite small. 
Machine learning models adapted to handle hyperspectral data such as \cite{sun2019spectral_1dconvs, hong2021spectralformer} were typically trained only on very small datasets. With our work we aim to train our proposed models on much larger and more diverse datasets, while using multiple tasks.

As has been identified by \cite{thompson_ml_promise_2021} and several overview papers \cite{signoroni2019deep, paoletti2019deep, gewali2018machine}, research in machine learning for processing hyperspectral data has been limited mainly due to the lack of large, publicly available datasets with high quality annotations. Many works still use very small datasets such as Indian Pines \cite{indian_pines_2015}, University of Pavia, Houston or Salinas, all of which contain only a few data captures. As further noted by \cite{signoroni2019deep}, this data is often divided into almost equal train and test subsets, which leads to model overfitting and results with very high accuracies almost regardless of the used model. 
In the domain of methane leak detection, only the recent works of \cite{STARCOP, kumar2023methanemapper} release \fixM{datasets created from a larger archive of full hyperspectral data captures provided by NASA}{R1Q7}. The work of \cite{STARCOP} furthermore refines the available labels.
The works of \cite{joyce_using_2022, jahan2024vgg_quant} propose machine learning models for detection and quantification of methane leak events in hyperspectral data, however the used datasets and code/model weights are not published.
In contrast, we publish all our used code and trained models and present the created datasets in a machine learning ready format.

Contribution of our work compared to the prior works is the focus on the low compute environment and the corresponding need for highly efficient models. We propose our models with this constraint in mind and we also test them on limited compute devices that serve as proxies to the computational environment on-board of satellites \cite{flordal2021spacecloud_dorbit, furano2020towards}.
We note that one could use off-the-shelf efficient machine learning models \cite{cai2023efficientvit_base} or apply methods such as pruning \cite{yang2023pruning_segformers} to existing machine learning architectures to improve the inference speed of the system. 
However, this would not address the problem of informational loss either due to the reliance on classical products, or due to informational bottlenecks encountered in not adapted architectures.

% // ML on-board of satellites
Deployment of machine leaning models on-board of satellites with hyperspectral sensors was shown in \cite{giuffrida2020cloudscout} to filter cloudy data captures.
The demonstrator Earth Observer 1 (EO-1) mission by NASA hosted simple thresholding models that processed hyperspectral data on-board to detect clouds and pre-classify scenes \cite{chien2005using_EO1autonomy} and later used models for mineral identification in \cite{thompson2012autonomous_on_EO1}.
On-board deployment of unsupervised models to detect anomalous events in multispectral data was proposed in \cite{RaVAEn} and later tested on-board of a real satellite in \cite{TrainOnBoard}.
These examples motivate us that methane detection using deep learning models on the current generation of hardware on today's satellites would be feasible. They also highlight the potential benefits from autonomy on-board in future satellite constellations.

\section{Data}
\label{sec:data}

\begin{figure}[!h]
  \centering
    % [trim={left bottom right top},clip]
  \subfloat[a][OxHyperSyntheticCH4, OxHyperMinerals]{\includegraphics[trim={0 0 0 2.0cm},clip,width=0.9\linewidth]{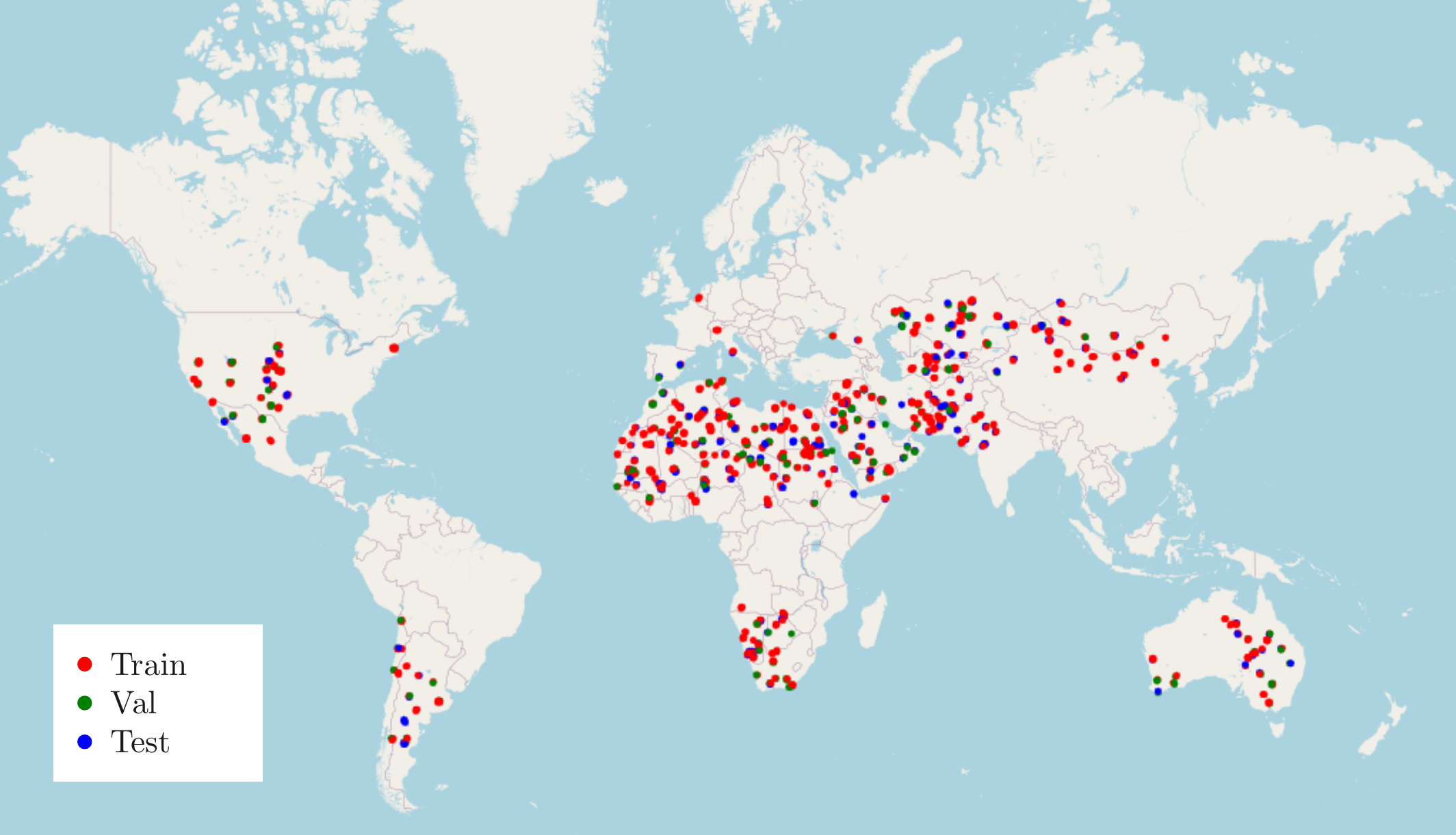}\label{fig:map_a}} \\
  \subfloat[b][OxHyperRealCH4]{\includegraphics[trim={0 0 0 2.0cm},clip,width=0.9\linewidth]{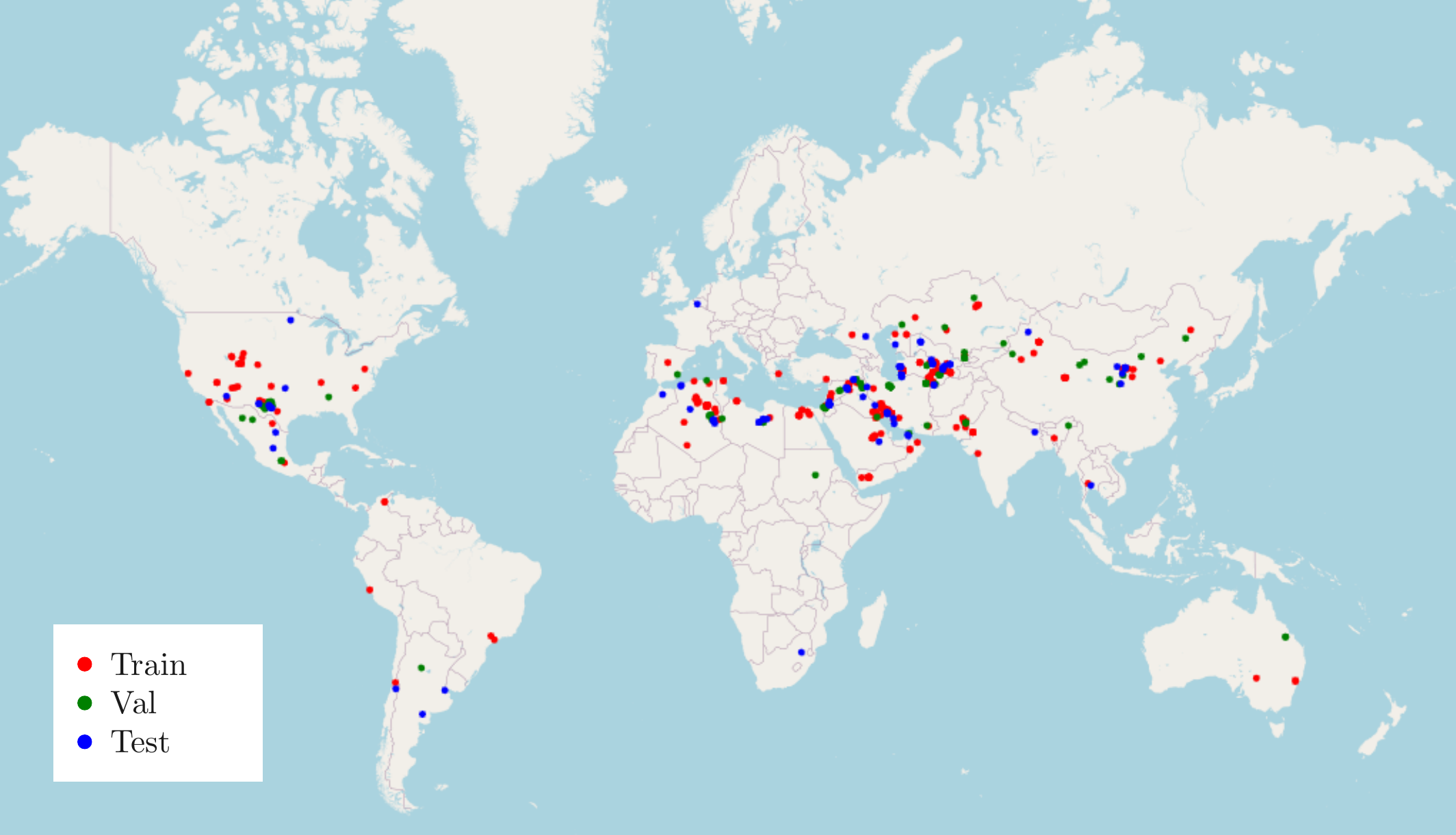}\label{fig:map_b}}
  \caption{Locations of tiles used to create the OxHyper datasets of EMIT data. We note that the same source tiles (including the same dataset splits) are used for the minerals (where they contain more spectral bands) and the synthetic methane datasets (where the methane leak events are added into the clean datacubes). \label{fig:hyper_maps}}
\end{figure}

In order to train machine learning models, which would be able to generalise globally around the world, we primarily need global and large datasets with high quality labels. As we already noted, many of the typically used hyperspectral datasets are very small. Recent methane leak datasets from \cite{STARCOP, kumar2023methanemapper} are large enough to train deep learning models, however they both use data from the aerial AVIRIS-NG sensor, which is relatively local in scope (e.g. the Four Corners region of USA). Instead, we use the data from the recent hyperspectral sensor Earth Surface Mineral Dust Source Investigation (EMIT), which is deployed on the International Space Station and has near global coverage of the arid regions of the Earth \cite{green2022_EMIT}.
In total we create three new datasets from EMIT data, their global coverage is demonstrated in Figure \ref{fig:hyper_maps} and they are described in detail in Table \ref{tab:tab_hyper_datasets}.

\subsection{Methane leak event simulation}

In one of our datasets, we simulate synthetic methane leak events in the clean hyperspectral datacubes, given the knowledge of the expected signal of methane and plume concentration data from real events. Unlike the previous study of \cite{jongaramrungruang_methanet_2022}, which used plume shapes simulated using the expansive Large Eddy Simulation (LES) \cite{matheou2014large_LES}, we use real methane leak events from the similar hyperspectral sensor AVIRIS-NG, to inform the simulation. Namely, we use the STARCOP dataset \cite{STARCOP}, given it’s improved and manually refined labels and high diversity of plume sizes.

Mathematically, we can describe the relation between a datacube with and without methane leak event using the Beer–Lambert absorption law \cite{foote_fast_2020_mag1c} as:

\begin{equation}
L_{simulated} ( \alpha, s ) = L_{clean} * e ^ {- \alpha * s}
\label{eq:simulation}
\end{equation}

Where $L$ refers to at sensor radiances in the hyperspectral datacubes (clean and resulting simulated one), $\alpha$ to the concentration of methane at given location and $s$ the methane absorption at given wavelength. We note that the quantity $\alpha$ is usually the one we are trying to estimate with matched filter approaches \cite{foote_fast_2020_mag1c}. As such, we can use existing matched filter products as sources for this quantity when simulating synthetic events. For $s$ we query the values of the methane signature at the wavelength of each band as plotted on Figure \ref{fig:methaneXsatellites}.

We then recompute the matched filter product from the simulated datacubes, this product now contains both the simulated event and the background profile of the scene (including confounder noise).

\subsection{Mineral aggregation}

\begin{figure}[h!]
    \centering
    \includegraphics[width=1.0\linewidth]{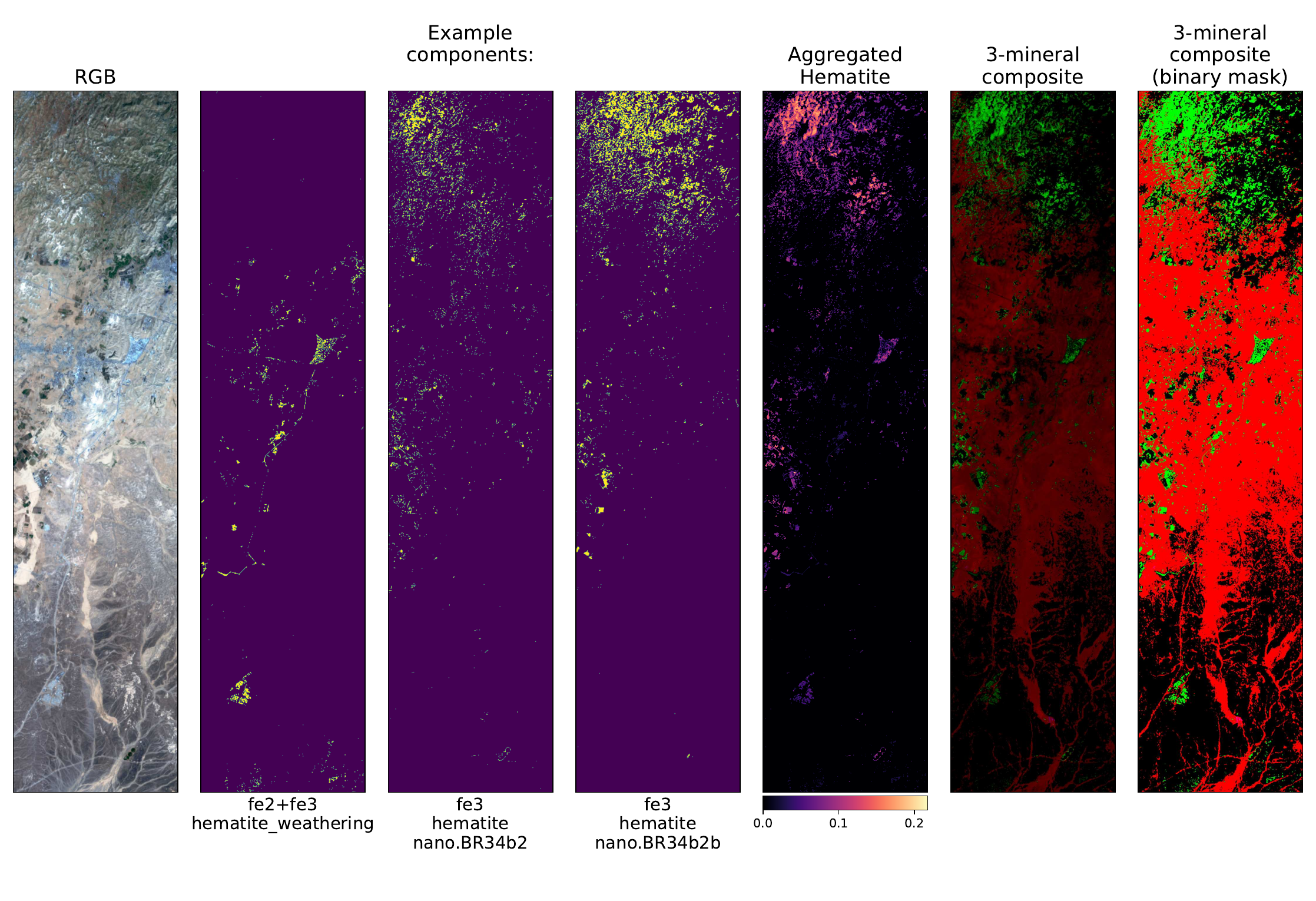}
    \caption{Aggregation of mineral components into mineral classes. 
    From left to right we show a RGB visualisation of the scene, binary mask of three common components for Hematite, the aggregated Hematite product (using in total 14 components) and visualisations of a 3-minerals composites. 
    We show the ``Goethite'', ``Hematite'' and ``Kaolinite'' composite and the binary maps used for training (in the last column).
    \label{fig:minerals_aggregation}
    }
\end{figure}

To create our new mineral dataset, we adapt the steps described in the {Theoretical basis document} \cite{ATBD_EMITL2B_v02} released alongside the ``EMITL2BMIN.001'' product \cite{DATA_EMITL2BMIN} computed from the EMIT data. This data contains identifications of 381 individual components (separated into two groups). These can be aggreggated into larger mineral classes, for this we select the so-called ``EMIT10'' subset defined in \cite{ATBD_EMITL2B_v02}. The aggregation process is shown on Figure \ref{fig:minerals_aggregation}. We note that we use only the binary maps of three selected mineral classes out of the total 10. We chose the subset M = \{Goethite, Hematite, Kaolinite\}, as these classes were abundant in the locations sampled in our dataset. We use the released annotations which were computed using the Tetracorder method \cite{clark2003imaging_Tetracorder}, and as such we can't really use Tetracorder as a classical baseline (as it would unfairly obtain perfect scores). Therefore, we consider the created dataset as having pseudo-ground truth labels, and our results only as initial proof-of-principle. We however note, that given any manually validated dataset, our models could be easily retrained with the new data.

\subsection{Final created datasets}

\begin{table}[h]
\caption{Overview of datasets created in this paper. We report the number of large tiles of 512x512 px, note that we usually further tile these into 64x64px tiles with 32px overlap.}
\label{tab:tab_hyper_datasets}
\centering
\scalebox{0.9}{
\begin{tabular}{@{}lrrrrrr@{}}
\toprule
\textbf{OxHyper-} & \textbf{Bands} & \textbf{Spectral range (nm)} & \textbf{Train}  & \textbf{Val} & \textbf{Test} \\ \midrule
SyntheticCH4 & 86 & 1573-1699, 2004-2478 & 796 & 198 & 200 \\
RealCH4 & 86 & 1573-1699, 2004-2478 & 279 & 91 & 98 \\
Minerals & 285 & 381-2492 & 796 & 198 & 200 \\ \midrule
STARCOP \cite{STARCOP} & 125 & RGB, 1573-1699, 2004-2480 & 3425 & N/A & 342 \\ \bottomrule
\end{tabular}
}
\end{table}

Firstly, we created the ``OxHyperSyntheticCH4'' dataset of synthetically simulated methane leak events using the labels of the STARCOP dataset with newly downloaded near-global hyperspectral datacubes from the EMIT sensor (in L1B level of data processing). In total we keep 796 large tiles (of 512x512 px) in the training dataset, 198 in the validation dataset and 200 in the test dataset. Each subset is made from completely non-overlapping sources of data (clean EMIT datacubes and labels from the STARCOP dataset from AVIRIS-NG data). In each subset we uniformly sample the available plume sizes, so that we have similar distribution of event sizes.
Half of these tiles contain simulated events while the other half is kept event free to provide negative samples.
We note that these large tiles (of 512x512 px) are further tiled when training the models. 
\fixM{All datasets created in this paper use tile size of 64 pixels with an overlap of 32 pixels. 
When evaluating the models (either on the Val or Test subsets), we keep the full resolution of 512x512 px.
The synthetic Train dataset gets tiled into 174341 samples.}{R1Q5}
The total size of this data is 228 GB.

Secondly, we create the ``OxHyperRealCH4'' dataset of real methane leak events. In total we have 279 large tiles in the training dataset, 91 in the validation dataset and 98 in the test dataset. 
When further tiled, the training dataset has 54805 samples.
These have reliable ground truth labels, which we calculated from the original labels provided on the EMIT data portal and consequentially manually checked. We note that this dataset is smaller than the other ones and training a machine learning model from scratch could lead to overfitting and poor generalisation, however it can also be used for model fine-tuning and to test generalisation ability of models trained on other datasets. The total size of this data is 47 GB.

Thirdly, we create ``OxHyperMinerals'', a completely new dataset for mineral identification, with labels of 3 selected mineral classes and source annotation for all 381 constituents (separated into two groups). In total we have 796 large tiles in the training dataset, 198 in the validation and finally 200 in the test dataset.
When further tiled, the training dataset has 174341 samples.
The large tiles exactly match the tiles used in the synthetic methane events dataset, however they do not contain the simulated events and contain more bands. The mineral dataset contains all 285 spectral bands in between spectral range of 381-2492nm.
The total size of this data is 372 GB. 

Finally, we also use a variant of the STARCOP dataset from \cite{STARCOP} with all relevant spectral bands extracted from the source data. 
\fixM{The STARCOP dataset uses tile size of 128 pixels with an overlap of 64 pixels for its Train subset. The Test subset is kept in its original resolution of 512x512 px.}{R1Q5}
We train our models and compare them on this benchmark dataset with the previously reported results.

\section{Methodology}
\label{sec:methodology}

\subsection{Limitations of existing approaches}

We are using three different semantic segmentation architectures with the created dataset of hyperspectral datacubes - the commonly used U-Net model \cite{ronneberger2015u} and the newer SegFormer \cite{xie2021segformer} and EfficientViT \cite{cai2023efficientvit_base} models. These models can be understood in terms of the encoder-decoder layout. The U-Net model consists of downsampling convolutional layers in the encoder network, and upsampling layers in the decoder, while also using skip-connections to communicate across the same spatial resolution of the data. The SegFormer model uses Transformer blocks in its encoder network, concatenates the features from different resolution levels and then uses a MLP block in the decoder network. The EfficientViT model further proposes changes to make the model run faster on limited hardware. In all models the spatial resolution is initially reduced and then reconstructed in the decoder, while information at different scales is kept with the use of skip-connections, or when concatenating the features together.

We are interested in training these architectures in an end-to-end manner on the \fix{full ranges of selected bands of} hyperspectral datacubes. In the methane detection task, this allows us to not depend on the classical matched filter product, which can be slow and which contains false detections of confounder materials. Models designed with typical computer vision data in mind (with one to three bands) however suffer from what we call informational bottleneck when they are used with data with large spectral dimensionality. As an example, if we select 86 bands of data which are relevant for our task of methane detection, the first layer in the vanilla versions of the used models reduces the data volume by $-95.34\%$ in the case of either the SegFormer, EfficientViT or U-Net (instead of increasing this volume by $+33\%$ as is common with RGB data). We claim that this informational bottleneck forces the model to get rid of most of the potentially valuable information right at the entry point to the model, which means that the later layers cannot effectively extract high level compressed information, as this extraction has to occur in the first convolutional layer. 

Furthermore, we note that the SegFormer and EfficientViT architectures reduce the spatial resolution of the input data by the factor of 4, and use non-parametric bilinear interpolation to scale the predicted output to the original resolution. In contrast, the U-Net architecture reduces the spatial dimension of the data only by the factor of 2, and uses a learned Upscaling layer at the end of the model to reconstruct the original resolution. 

\subsection{Proposed architecture}

\begin{figure*}
    \centering
    \includegraphics[width=0.85\linewidth]{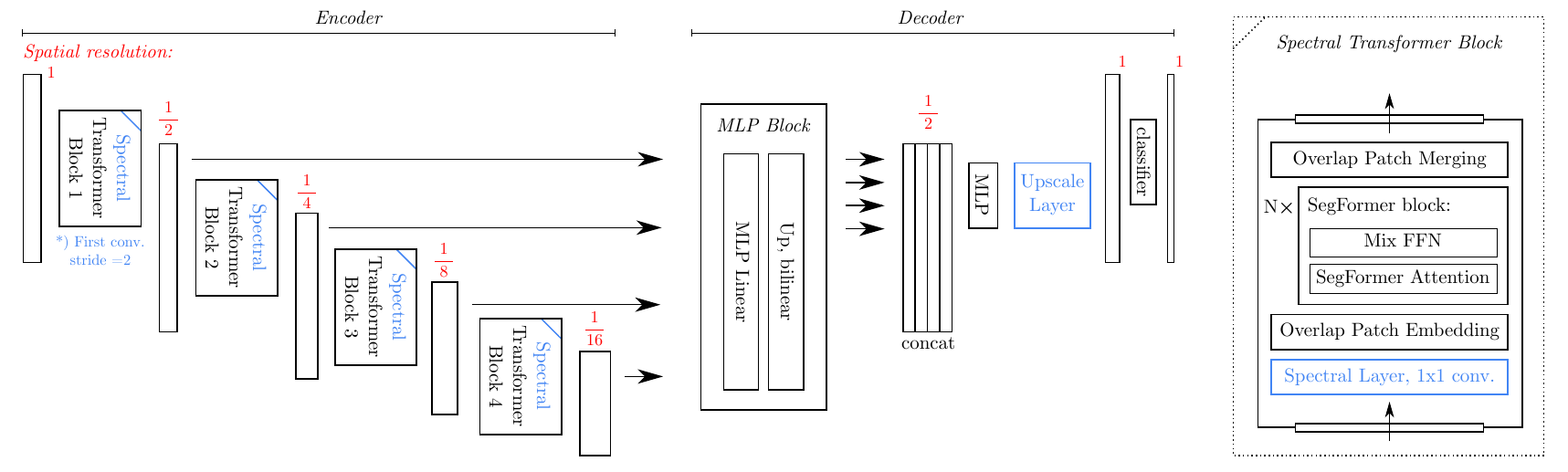}
    \caption{Illustration of the Hyper SegFormer model for semantic segmentation of methane plumes in Hyperspectral data. We highlight in blue the three proposed modular adjustments: 1.) Spectral layers (denoted as ``1x1 Conv'') in the Transformer blocks, 2.) Upscaling layer in the decoder network (denoted as ``Upscale layer'') and finally 3.) adjusting the stride of the first Transformer block to 2 (denoted as ``Stride''). In red we show an example of the progressive decimation of the resolution throughout the model (using all three adjustments) - we highlight that the typical off-the-shelf variants of the models reduce the output resolution by a factor of 4.}
    \label{fig:spectral_segformer_model}
\vspace{-2mm}%less space in between figs 
\end{figure*}

We propose three modular changes to the selected architectures to allow for better handling of data with high spectral dimension (such as data from hyperspectral satellites) as illustrated on Figure \ref{fig:spectral_segformer_model}. 

Firstly, to aid with the spectral dimensionality of the data, we propose an addition of a special spectral layer into all Transformer blocks throughout the encoder networks. We add a Convolutional layer with kernel size of (1,1), stride of 1 and padding of 0, keeping the output of each layer with the same spectral dimension as on its input. This allows for learning operations that occur merely on the spectral data, on its own these allude to pixel-by-pixel operations of classical methods. While in theory the original convolutional layers with larger kernel sizes could learn similar operations, by restricting the operation to just the 1x1 convolutions, we obtain much sharper outputs. Effectively, this also allows the model to weight different input bands with a mechanism similar to attention, thus selecting only the relevant information for the task of interest. While this is a simple addition to the model, we note that it is quite effective, and it doesn’t significantly increase the computational requirements of the model.

Secondly, we address the reduction of the spatial resolution of the data. We implement a similar Upscaling layer to the one used in the U-Net architecture at the end of the Transformer based models, just before their classification heads. 
This block is composed out of two repeated 2D convolutional layers with kernel size of (3,3), padding and stride of 1, each followed by a ReLu activation layer and Batch normalisation layer. 

As an additional solution for the recovery of the lost resolution, we also explore the change to the hyperparameters of the original architectures, namely changing the stride of the first convolutional layer (e.g. in SegFormer from 4 to 2). This change effectively increases the spatial size reduction of the data propagated through the network by the factor of 2. This doesn’t change the number of parameters of the resulting model, but as the effect of this change influences the whole model, the impact on the inference speed is larger than with the Upscaling layer.

These three proposed changes are modular and can be evaluated in isolation or used all together – the resolution of the output product will either be ¼, ½, or 1 times the resolution of the input. 

\subsection{Baselines}

As a classical baseline approach for methane detection in hyperspectral data, we use the matched filter product (using the iterative variant mag1c \cite{foote_fast_2020_mag1c}) with the threshold of 500 ppm$\times$m. Additionally, we apply an opening morphological filter to remove isolated noisy pixels in the product.
\fixM{More specifically, we perform the erosion and dilation operations (using the implementations from the Kornia library). We explore two kernel shapes: ``cross'' kernel (3x3 pixels matrix with the middle row and column full of ones) and ``ones'' kernel (3x3 pixels matrix with all values set to ones). We observed small differences when trying these kernels, the ``cross'' kernel further corresponds to the method used in \cite{STARCOP}. We observed that other kernel shapes, larger sizes, or different initial thresholds didn't improve the results.}{R1Q2}

As a classical machine learning baseline, we implement the HyperSTARCOP model based on the U-Net architecture proposed by \cite{STARCOP} and train it on our new methane datasets. This model uses the pre-computed matched filter product alongside the RGB bands of the input data and effectively learns to clean the classical methane enhancement product. We note, that if the matched filter product misses a methane leak event, this model has no way of accessing the original information in the hyperspectral datacube to recover the detection.

For the task of mineral identification, we can't use the classical method of Tetracorder as a baseline, as we used it to generate our pseudo-ground truth labels. When inspecting other baseline approaches such as for example the spectral angle mapper (SAM)  \cite{yuhas1992discrimination_SAM}, we noted significant disagreement between the two methods. As such, for the task of mineral identification, we chose to only compare different versions of our adapted models against their default out-of-the-shelf version which we treat as a deep learning baseline.

\section{Experimental Setup}

When training machine learning models on our newly created datasets, we use the following settings and hyperparameters.
We train our models on the task of semantic segmentation, in methane detection with two exclusive classes (``plume'' and ``no plume'') and in mineral detection with a multi-hot label (which allows no detection or potentially multiple active classes of ``Goethite'', ``Hematite'' and ``Kaolinite''). 

During training on the methane datasets, to balance the data distribution, we oversample the instances with plume pixels by using the WeightedRandomSampler provided by the PyTorch library. 
We use the Adam optimiser with learning rate of 0.001 for the U-Net architectures and 0.00006 for the SegFormer and EfficientViT architectures. 
U-Net models use the MobileNet-v2 backbone \cite{sandler2018mobilenetv2} in the encoder, SegFormer uses the B0 backbone architecture and EfficientViT the B1 backbone architecture.
We use the batch size of 32 with U-Net models and 16 with SegFormer models, due to memory limitations while training. 

On the ``OxHyperSyntheticCH4'' dataset we train all models for 50 epochs. When training we use the loss on the set aside validation subset to select the best performing models.
When fine-tuning the models on the ``OxHyperRealCH4'' dataset, we train the models for further 10 epochs from the best reached checkpoint. When training models from scratch on this smaller dataset, we keep the number of epochs to 50. End-to-end models use 86 bands of the EMIT data (RGB and bands between 1573-1699nm and 2004-2478nm).

To compare the newly proposed models with the HyperSTARCOP ``RGB+MF'' model from \cite{STARCOP}, we also train them on the AVIRIS-NG STARCOP dataset. Here we train the models for 15 epochs using a batch size of 16 and tile size of 128 px with an overlap of 64 px. Except for the batch size (original was 32), these hyperparameters are kept the same as in the original paper. We also weight the model loss by the MF product, as was done with the original HyperSTARCOP models. 
As there is no validation subset, we use the model from the last epoch.
All end-to-end models use 60 bands of AVIRIS-NG (namely bands between 2104-2400 nm).
This was chosen to match the range typically used by the matched filter as shown in Figure \ref{fig:methaneXsatellites}.

Finally, when training models on the ``OxHyperMinerals'' dataset on the task of mineral detection, we train all models for 20 epochs with tiles of 64 px with 32 px overlap. In this case, trained models use all 285 bands of the available EMIT data (381-2492 nm). Here, the data loader is not balancing the samples and losses are not weighted.

\subsection{Hardware Restrictions}
\label{sec:hw_res}

For training, we use a high performance computing cluster with Tesla V100 GPUs with 32GB GPU memory and the Intel Platinum 8628 (Cascade Lake), 2.90GHz CPU with 384GB RAM. 
This type of machine is typically much more computationally capable than the hardware later used for deployment.

Therefore, with several tested model variants, we also conducted experiments measuring the runtime on compute constrained hardware, namely on the flatbed hardware provided by Unibap and D-Orbit. 
These machines serve as a proxy to match the computational in-space environment of a D-Orbit's ION-SCV004 satellite, which is equipped with: quad-core 1.2GHz, AMD GX-412HC SOC CPU processor, the Intel Movidius Myriad X VPU and 2GB of RAM.
\fixM{The power consumption of the used environment (SpaceCloud® iX5-100), is estimated to about 18W by \cite{flordal2021spacecloud_dorbit}.}{R2Q2}
The same compute environment was used in \cite{TrainOnBoard} and similar environments leveraging Myriad chips were used on prior missions such as ESA's \fixM{Phi-Sat-1}{R1Q9} \cite{giuffrida2021phi_sat_1_mission} and \fix{Phi-Sat-2} \cite{marin2021phi_sat2}.

We have also selected two other devices typically used for Internet-of-Things (IoT) applications, the CPU powered Raspberry Pi boards and the GPU equipped NVIDIA Jetson AGX Xavier board.
As has been noted by NASA's guidelines on using Raspberry Pi's in space \cite{guertin2022raspberry}, they are not suggested for deployment in any mission critical tasks due to their lower robustness in extreme environments. 
They have been however used as a proxies for devices of similar compute power in \cite{gretok2021onboard_pynq_zynq}.
\fixM{The estimated power consumption of Raspberry Pi devices is around 5W during a benchmark.}{R2Q2}
We then use the more powerful NVIDIA Jetson AGX Xavier board equipped with an 8-core NVIDIA Carmel ARMv8.2 CPU at 2.265 GHz and a GV10B GPU based on the Volta architecture with 16GB memory (shared between CPU and GPU).
\fixM{The power consumption of this board can be adjusted, in our case we used the 30W mode.}{R2Q2}
We used both ONNX Runtime python libraries and the Tensor RT inference engine (using the \textit{trtexec} command), as was also used in the work of \cite{cai2023efficientvit_base}.
While initial tests of the effects of radiation on NVIDIA's Jetson devices was explored in \cite{slater2020total} and proposed for space applications in \cite{lofqvist2020accelerating}, these works also state that a careful examination and future tests are still needed.

\section{Results}
\label{sec:results}

\begin{table*}[]
\caption{Quantitative results on the OxHyperSyntheticCH4 dataset. We show the average of training 5 runs of our models. We also compare the results against the HyperSTARCOP model from \cite{STARCOP}.} \label{tab:emit_synth_quant}
\centering
\scalebox{1.0}{
\begin{tabular}{@{}lllllll@{}}
\toprule
                            & AUPRC ↑              & F1 ↑                  & Precision ↑           & Recall ↑              & IoU ↑        & FPR by tile ↓          \\ \midrule
\fix{Baseline, mag1c, morpho. (cross 3x3)}   & N/A                  & 37.65                & 31.69                & 46.37                & 23.19                & 89.0                \\
\fix{Baseline, mag1c, morpho. (ones 3x3)}   & \fix{N/A}                  & \fix{41.46}                & \fix{41.68}                & \fix{41.25}                & \fix{26.15}                & \fix{79.0}                \\
HyperSTARCOP MF+RGB \cite{STARCOP} & 68.10 ± 4.7          & 58.08 ± 5.3          & 48.44 ± 6.7          & 73.31 ± 2.8          & 41.12 ± 5.2          & 65.6 ± 6.7          \\ \midrule
SegFormer base              & 67.76 ± 2.6          & 60.26 ± 1.7          & 53.14 ± 4.9          & 70.70 ± 5.8          & 43.15 ± 1.7          & 27.2 ± 3.1          \\
SegFormer ConvUp            & 78.14 ± 3.0          & 71.02 ± 3.4          & 68.72 ± 7.4          & 74.17 ± 3.2          & 55.18 ± 4.2          & 25.2 ± 7.2          \\
SegFormer ConvUpStride      & \textbf{82.24 ± 2.7} & \textbf{74.27 ± 2.9} & \textbf{74.15 ± 4.0} & \textbf{74.88 ± 6.0} & \textbf{59.15 ± 3.7} & \textbf{18.4 ± 2.4} \\ \midrule
EfficientViT base           & 74.29 ± 7.3          & 68.40 ± 6.9          & 65.54 ± 10.3         & 72.75 ± 6.0          & 52.39 ± 7.8          & 25.0 ± 5.0          \\
EfficientViT ConvUp         & 77.86 ± 3.4          & 70.89 ± 3.1          & 68.31 ± 5.8          & 74.06 ± 3.1          & 55.00 ± 3.7          & 40.2 ± 9.2          \\
EfficientViT ConvUpStride   & 78.94 ± 5.3          & 72.26 ± 4.9          & 72.63 ± 6.3          & 72.19 ± 5.5          & 56.80 ± 6.0          & 31.8 ± 5.9          \\ \bottomrule
\end{tabular}
}
\end{table*}

In this section, we first compare the results from different used models on the task of methane detection in hyperspectral data from two explored sensors.
Secondly, we report the results from using our proposed models on a novel dataset for mineral identification.
Finally, we also report measurements of the inference time needed to use these models on a compute constrained hardware of the Unibap flatsats. These will serve as an approximate proxy to the runtime needed on-board of potential satellites.

\subsection{Methane detection on EMIT}

\begin{figure}[!h]
    \centering
    \includegraphics[width=0.95\linewidth]{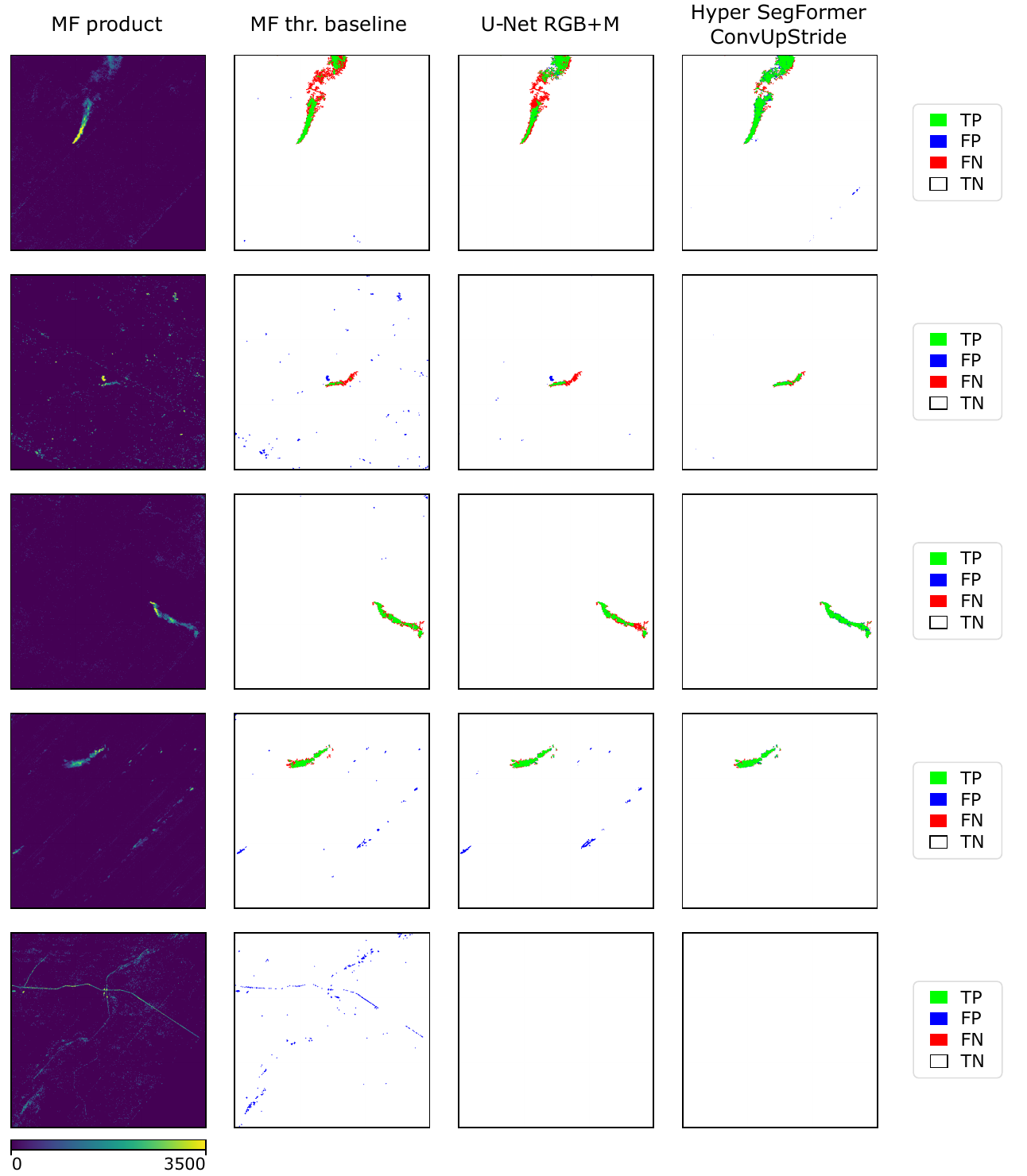}
    \caption{Qualitative results on the OxHyperSyntheticCH4 dataset showing the computed matched filter in the first column and comparing the performance of the matched filter baseline, U-Net based model from \cite{STARCOP} and our proposed Hyper SegFormer model (``ConvUpStride'' variant).}
    \label{fig:emit_synth_qualitative}
\vspace{-2mm}%less space in between figs 
\end{figure}

Figure \ref{fig:emit_synth_qualitative} shows the comparison on the selected methods - classical baseline, U-Net based model from \cite{STARCOP} and our proposed Hyper SegFormer model (using the ``1x1 Conv, Up, Stride'' variant). We illustrate the advantage of our proposed solution. The baseline and the U-Net based models depend on the matched filter product (also shown in the first column), and as such, if this product doesn't capture the full extent of the methane plume present in the data, these models aren't able to correctly reconstruct it. Meanwhile, our proposed model has access to the original source data and can extract information which would be lost when using the matched filter product. We show this in the first four rows. The last row demonstrates that if the matched filter product contains false detections (confounder materials such as roads, and building outlines), both learned models can filter these out.

Table \ref{tab:emit_synth_quant} shows the results of several tested model variants for both SegFormer and EfficientViT architectures compared with classical machine learning approach of \cite{STARCOP} and baseline method using \cite{foote_fast_2020_mag1c} \fixM{(regardless of the choice of kernel used in the baselines, although the ``ones'' baseline variant improves most of the scores over ``cross'' originally used in \cite{STARCOP})}{R1Q2}. 
% We can describe the first two, depending on classical enhancement products, as ``few bands'' and the rest using end-to-end learning as ``many bands''.
We note that all of the deep learning approaches significantly outperform the classical baseline.
Furthermore, our proposed HyperSegFormer (ConvUpStride variant) gains over 43.85\% in the IoU score (and 27.88\% in F1 and 20.7\% in AUPRC) over the HyperSTARCOP MF+RGB model. Our model also achieves over 155\% in IoU score (and 96.9\% in F1 score) over the classical matched filter baseline. These improvements hold over all measured metrics. We see increasing scores when implementing proposed adaptations in both the SegFormer (from 60.26 F1 score to 74.27) and EfficientViT (from 68.40 F1 score to 72.26) of our HyperspectralViT models.

\begin{table*}[]
\caption{Quantitative results on the benchmark STARCOP dataset. We show the average of training 5 runs of our models.} \label{tab:aviris_compare_quant}
\centering
\scalebox{0.95}{
\begin{tabular}{@{}lcccccccc@{}}
\toprule
                            & AUPRC ↑                & F1 ↑                  & F1 (strong) ↑         & F1 (weak) ↑           & Precision ↑            & Recall ↑              & IoU ↑                 & FPR by tile ↓         \\ \midrule
\fix{Baseline, mag1c, morpho. (cross 3x3)}   & N/A                  & 40.14                & 67.50                & 39.95                & 30.57                & 58.42                & 25.11                & 75.43                \\
\fix{Baseline, mag1c, morpho. (ones 3x3)}   & \fix{N/A}                  & \fix{45.01}                & \fix{64.93}                & \fix{43.06}                & \fix{40.32}                & \fix{50.94}                & \fix{29.04}                & \fix{57.71}                \\
HyperSTARCOP MF+RGB \cite{STARCOP} & 51.99 ± 2.8          & 50.26 ± 3.8          & \textbf{81.96 ± 3.7} & 43.42 ± 5.7          & 37.53 ± 5.8          & \textbf{78.39 ± 6.6} & 33.63 ± 3.4          & 43.66 ± 7.4          \\ \midrule
SegFormer base              & 48.84 ± 3.3          & 37.88 ± 6.5          & 68.20 ± 3.2          & 17.91 ± 4.7          & 25.71 ± 6.3          & 75.58 ± 4.0          & 23.52 ± 5.0          & 32.80 ± 12.5         \\
SegFormer ConvUp            & 57.91 ± 5.6          & 53.17 ± 6.5          & 71.53 ± 4.0          & 39.37 ± 8.7          & 44.93 ± 10.8         & 68.16 ± 5.9          & 36.42 ± 6.0          & 32.34 ± 20.2         \\
SegFormer ConvUpStride      & \textbf{60.98 ± 4.5} & \textbf{56.75 ± 3.7} & 73.16 ± 1.9          & \textbf{45.65 ± 3.0} & \textbf{50.87 ± 7.7} & 65.73 ± 5.8          & \textbf{39.69 ± 3.5} & \textbf{19.89 ± 6.1} \\ \bottomrule
\end{tabular}
}
\end{table*}

\begin{figure}[!h]
  \centering
  \includegraphics[width=0.95\linewidth]{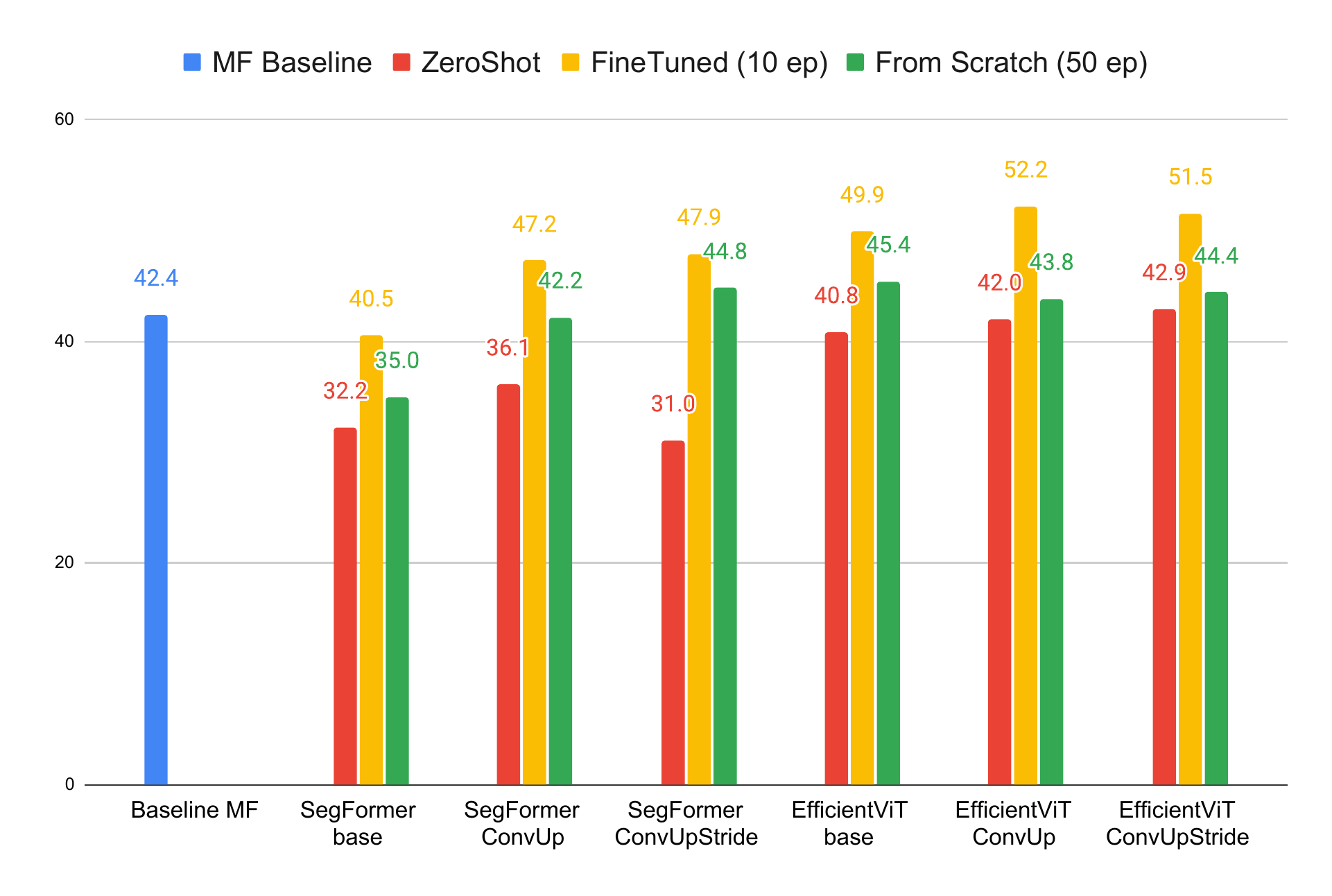}
  \caption{Quantitative results on the OxHyperRealCH4 dataset showing the F1 score. We compare the MF baseline against zero-shot generalisation, fine-tuning or training from scratch for both proposed models. Results are averaged from 5 training runs, and the values on top of individual bars are rounded for better clarity. \label{fig:emit_real_quant}}
\end{figure}

% 31, 52, 97, 77, 20
\begin{figure}[!h]
    \centering
    \includegraphics[width=0.95\linewidth]{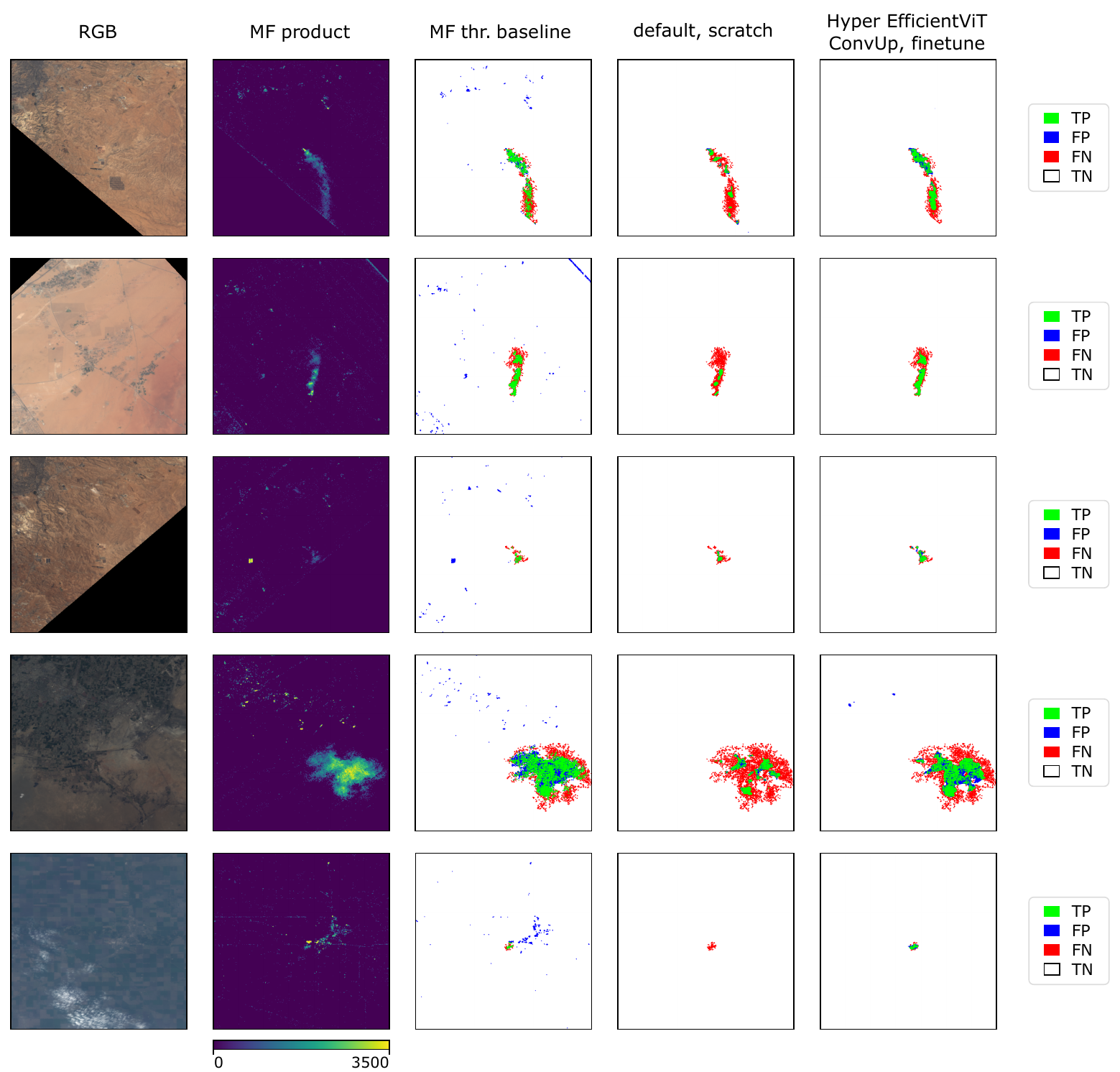}
    \caption{Qualitative results on the OxHyperRealCH4 dataset. From left to right we show the predictions of the classical baselines, models trained from scratch and models pre-trained on the synthetic dataset (using the HyperEfficientViT model, ConvUp variant). We note that black parts are the no-data areas created upon orthorectification.
    }
    \label{fig:emit_real_qualitative}
\vspace{-2mm}%less space in between figs 
\end{figure}

In Figure \ref{fig:emit_real_quant} we evaluate our adapted SegFormer and EfficientViT models on the OxHyperRealCH4 dataset of real methane leak events captured by the EMIT sensor. Figure \ref{fig:emit_real_qualitative} shows the qualitative results on this dataset. Here, we use the models pre-trained on the synthetic dataset (which uses data of the same input/output shapes and uses the same data normalisation) and test their zero-shot generalisation on the dataset of real events. Furthermore, we fine-tune these models on the smaller dataset of real events. Finally, we also compare these results with training models from scratch only on the dataset of real events. We note that this experiment resembles a real-world scenario, where we want to train models for a newly deployed sensor, which hasn't yet seen enough of these rare events. We try to answer the question of if pre-training on a synthetically created dataset helps the final models detect events on a real but small dataset.

Interestingly, when evaluating the models pre-trained on the dataset with synthetic events on a new dataset of real methane leak events collected by EMIT, we observe that the zero-shot generalisation performance is low (only two model variants reach the MF baseline \fix{- the ``cross'' kernel variant}). We however see that with fine-tuning for 10 epochs, we are able significantly outperform both the classical baselines and all model variants trained from scratch. 
Namely the HyperSegFormer (ConvUpStride variant) improves the performance of the MF baseline by 12.9\% in F1 score (and 17.1\% in IoU) and the same version of the model trained from scratch by over 6.9\% in F1 score (and 8.8\% in IoU). 
The HyperEfficientViT (ConvUp variant) outperforms the MF baselines by more than 23\% in F1 (31.5\% in IoU) and the same model variant trained from scratch by 19.2\% in F1 (and 25.6\% in IoU). 
Interestingly, the ConvUp variant slightly outperforms the ConvUpStride variants on this dataset with the EfficientViT models. Overall, we see, that on the task of detecting real methane leak events in EMIT data, we benefit from pre-training our models on a dataset of synthetically created events (especially in cases similar to ours, where the dataset of real events is relatively small).

\subsection{Methane detection on AVIRIS-NG}

In Table \ref{tab:aviris_compare_quant} we evaluate our newly proposed end-to-end models on the benchmark dataset STARCOP \cite{STARCOP} of real methane leak events in data from the aerial AVIRIS-NG. We train our proposed end-to-end models from scratch matching the settings used in the original work of \cite{STARCOP}.

When training our proposed models on the previously created benchmark dataset of real methane leak events in AVIRIS-NG data, we see some interesting mixed outcomes. While the performance of our models drops on F1 score of strong methane leak events (plume events larger than 1000 px) in contrast to the HyperSTARCOP ``MF+RGB'' model, on the metrics that aggregate all event sizes together, we see improvements. Namely, the HypeSegFormer (ConvUpStride variant) outperforms the HyperSTARCOP ``MF+RGB'' baseline by 12.9\% in F1 score and by 18\% in IoU score. The threshold-less AUPRC metric is also improved by over 17\%. The high performance of previous STARCOP models on strong methane leak events could be explained by the fact that the used matched filter method, mag1c, works better for large plumes (as it is iteratively increasing the contrast of positive pixels - this could however lead to lowered performance for weaker events). 

We have encountered very large gains on the synthetic and real EMIT datasets, however smaller ones on the benchmark dataset of AVIRIS-NG data. We note that perhaps better versions of the MF product could lead to stronger baselines also in the EMIT data. For example \cite{foote2021impact_scene_specific} explores scene specific adaptation of MF products, and similarly there could be some EMIT specific improvements for calculating MF.
This difference could also be caused by the number of used bands - while on the STARCOP dataset the end-to-end models use 60 bands, on the new EMIT datasets we use 86 bands which also have a wider spectral range coverage.

\subsection{Mineral identification}

% RUNS06
\begin{table*}[]
\caption{Quantitative results on the mineral dataset OxHyperMinerals. We show the average of training 5 runs of our models.} \label{tab:emit_minerals_quant}
\centering
\scalebox{1.0}{
\begin{tabular}{@{}lccc|ccc@{}}
\toprule
                          & F1                    & Precision             & Recall                                     & F1 Goethite           & F1 Hematite           & F1 Kaolinite          \\ \midrule
SegFormer base            & 81.77 ± 1.54          & 77.33 ± 3.58          & \multicolumn{1}{c|}{\textbf{86.92 ± 1.79}} & 79.31 ± 1.08          & 80.88 ± 0.66          & 85.79 ± 3.84          \\
SegFormer ConvUp          & 82.99 ± 0.92          & 80.55 ± 1.01          & \multicolumn{1}{c|}{85.64 ± 2.26}          & 79.36 ± 1.55          & 81.47 ± 1.82          & 89.15 ± 0.89          \\
SegFormer ConvUpStride    & \textbf{84.70 ± 0.70} & 83.42 ± 2.69          & \multicolumn{1}{c|}{86.14 ± 1.98}          & \textbf{81.19 ± 1.03} & \textbf{83.59 ± 1.41} & \textbf{90.35 ± 0.80} \\ \midrule
EfficientViT base         & 77.08 ± 0.74          & 79.37 ± 3.73          & \multicolumn{1}{c|}{75.28 ± 3.99}          & 69.88 ± 3.09          & 77.89 ± 1.09          & 85.25 ± 1.05          \\
EfficientViT ConvUp       & 79.73 ± 0.92          & 82.66 ± 2.32          & \multicolumn{1}{c|}{77.11 ± 2.39}          & 74.59 ± 1.28          & 79.74 ± 0.67          & 86.55 ± 1.30          \\
EfficientViT ConvUpStride & 79.18 ± 1.16          & \textbf{84.03 ± 2.74} & \multicolumn{1}{c|}{74.96 ± 2.17}          & 73.57 ± 1.20          & 77.93 ± 2.70          & 87.24 ± 1.22          \\ \bottomrule
\end{tabular}
}
\end{table*}

\begin{figure}[!h]
    \centering 
    \includegraphics[width=0.9\linewidth]{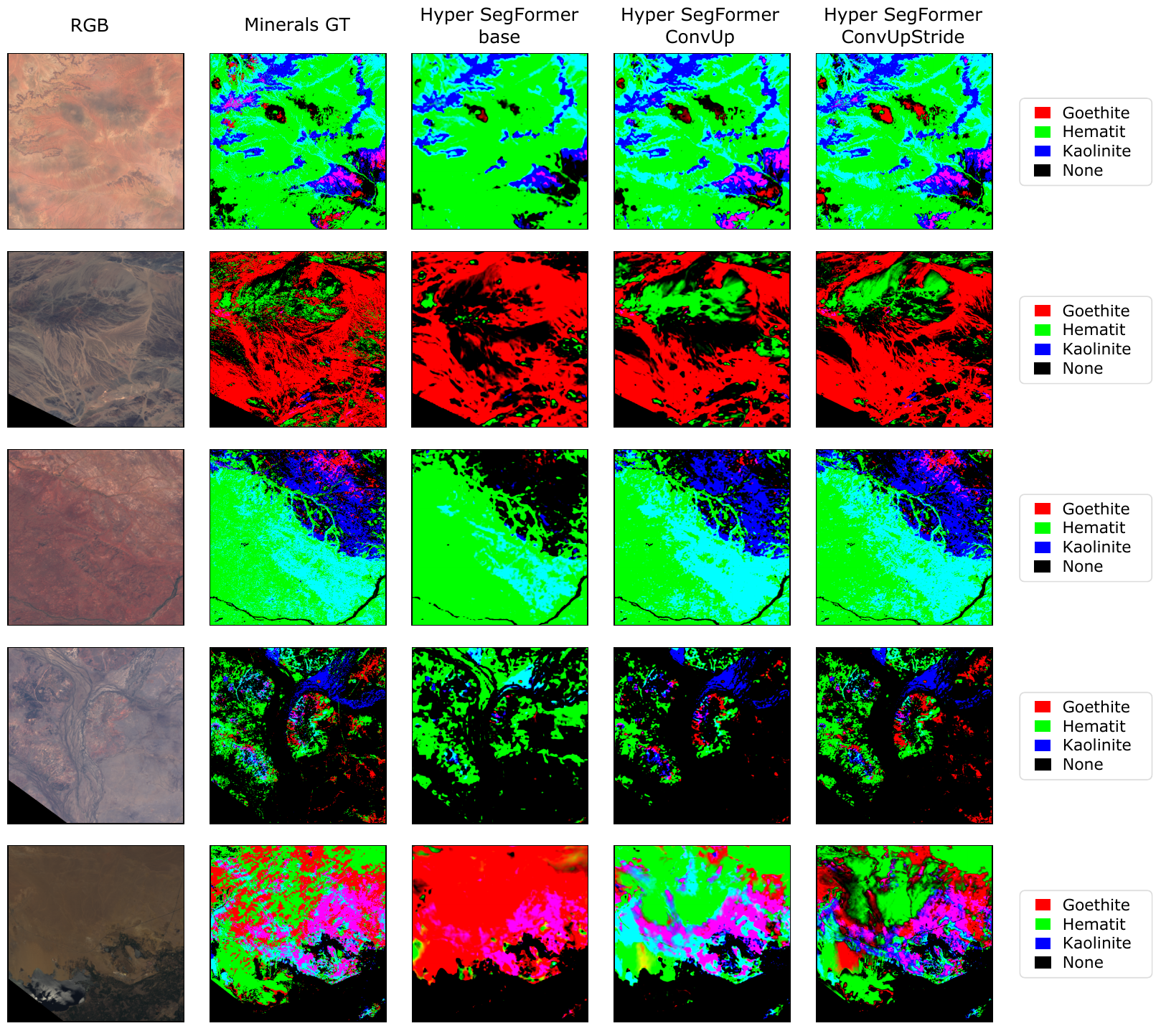}
    \caption{Qualitative results comparing our new models on the mineral dataset OxHyperMinerals showing our proposed Hyper SegFormer model variants (with output resolution degradation in default ¼x, ``ConvUp'' ½x, and ``ConvUpStride'' full 1x). 
    % The three mineral classes ``Goethite'', ``Hematite'' and ``Kaolinite'' are mapped onto as the ``Red'', ``Blue'' and ``Green'' channels for visualisation, 
    % Model output is not thresholded and matches the saturation.
    }
    \label{fig:emit_minerals_qualitative}
\vspace{-2mm}%less space in between figs 
\end{figure}

In Table \ref{tab:emit_minerals_quant}, we show the results of training the proposed machine learning architectures on the mineral dataset OxHyperMinerals. We show the individual F1 scores for each of the three mineral classes alongside the aggregated scores (which are support weighted by the number of pixels belonging to each class). In Figure \ref{fig:emit_minerals_qualitative} we also show examples of model predictions next to the pseudo-ground truth generated by the Tetracorder system. Note the gradual increase of output resolution depending on the used adaptations. 

When exploring the results of models trained on the dataset of minerals, we note smaller but sustained gains in performance due to the proposed adaptations. Namely the HyperSegFormer (ConvUpStride variant) improves the F1 score by over 3.5\% in contrast to the vanilla model (and precision by 7.9\%). The HyperEfficientViT demonstrates weaker performance for both base and adapted versions, however the used adaptations do improve this performance - the ConvUp variant scores 3.4\% improvement in the F1 score.

Perhaps better improvement can be seen on the qualitative comparison between the models on Figure \ref{fig:emit_minerals_qualitative}, where we show the effect of gradually improving output resolution with the HyperSegFormer model. The model variants using ``Up'' and ``Stride'' adaptations (and their combination) show sharper predictions than the vanilla version of the model (as that one is limited to the output size reduced by the factor of 4).

We note that our experiments with mineral identification are initial, as we had to rely on pseudo-ground truth labels generated by Tetracorder.

\subsection{Inference times on low compute hardware}

\begin{table*}[]
\caption{Timing of models on different devices. We estimate the time in seconds for processing a full granule of EMIT data (1280x1242px tiled into 100 tiles of 128x128px). EfficientViT models weren't evaluated on the Unibap machine. IO time to load the hyperspectral tiles is kept separate in the last row. We also illustrate the number of parameters each model uses.} \label{tab:timings_granule}
\centering
\scalebox{1.0}{
% \begin{tabular}{@{}lccccc@{}}
\begin{tabular}{@{}l|c|c|ccc|cc@{}}
\toprule
                          & Params. (M)  & Unibap CPU & Rasp3 CPU & Rasp4 CPU & Xavier CPU & X. GPU (onnx) & X. GPU (trtexec) \\ \midrule
HyperSTARCOP MF+RGB \cite{STARCOP}   & 6.633    & 203.3      & 1025.25   & 286.57    & 377        & 371.17            & 370.27               \\ \midrule
SegFormer Base           & 3.845  & 15.3       & 38.82     & 9.19      & 7.87       & 1.62              & 0.49                 \\
SegFormer UpConv         & 4.326  & 30.1       & 105.43    & 28.7      & 13.92      & 1.97              & 0.64                 \\
SegFormer UpConvStride   & 4.326  & 119.5      & 399.76    & 107.87    & 48.38      & 2.95              & 1.37                 \\ \midrule
EfficientViT Base        & 4.811  & N/A        & 39.96     & 9.45      & 9.39       & 2.36              & 0.37                 \\
EfficientViT ConvUp       & 4.855 & N/A        & 47.73     & 13.81     & 9.99       & 2.00              & 0.55                 \\
EfficientViT ConvUpStride  & 4.855 & N/A        & 91.07     & 21.26     & 18.09      & 2.10              & 0.5                  \\ \midrule
IO                       & N/A  & 4          & 39.99     & 12.88     & 6.13       & 6.13              & 6.13                 \\ \bottomrule
\end{tabular}
}
\end{table*}

Table \ref{tab:timings_granule} shows the model inference times needed to process a full EMIT scene capture of 1280x1242 px with our models for methane leak detection. 
\fixM{We tile this full granule into 100 non-overlapping tiles (padding the data if needed) - we note that in this way we could circumvent limitations of fixed input sizes encountered for example with the Myriad X VPU chip.}{R1Q3}
These measurements show the IO time needed to load the hyperspectral datacube and the model inference time - for baselines this also includes the time needed to calculate the MF product. 

We note that the matched filter computation is the slowest step influencing both the classical baselines and the HyperSTARCOP models. Our proposed end-to-end model variants avoid this step gaining vastly reduced inference time. Namely, our proposed HyperSegFormer ConvUp variant reduces the required time by over 85\%, while the ConvUpStride variant reduces the time by over 41\% - both compared against the HyperSTARCOP model and effectively also the classical MF baseline.

When considering other devices, the MF calculation remains to be the bottleneck and the end-to-end models are faster. While using the CPU of the Raspberry Pi 3B+ the runtime is slower than on the Unibap flatbed, Raspberry Pi 4B reaches similar performance. Finally, the most powerful CPU of the NVIDIA's Jetson AGX Xavier board shows the fastest runtimes of the tested CPU devices.

If we use our proposed HyperSegFormer ConvUp model on the Jetson AGX Xavier GPU, we achieve over 47x times speed-up over the Unibap's machine CPU (e.g. 0.64 instead of 30.1 seconds per capture) and 317x times speed-up over the HyperSTARCOP model (203.3 seconds per capture).
Using the GPU of the Jetson AGX Xavier board provides the fastest runtimes from all tested devices. We observed further speedup when using the Tensor RT engine (using the ``trtexec'' tool) and compiling the neural network directly on the device with allowed post-training mixed precision quantization (allowing float32 and float16) - in the case of the HyperSegFormer ConvUp, this is by over 70\% in contrast to using the Onnxruntime library (without compilation and without mixed precision mode). 

When using the Tensor RT inference engine, we also experience faster inference times when using the adapted EfficientViT architecture instead of the adapted SegFormer architecture - this is consistent with the fact the EfficientViT was optimised for fast inference speed in \cite{cai2023efficientvit_base}. Namely for the ConvUpStride variants, the HyperEfficientViT has runtime lower by about 64\% in contrast with the HyperSegFormer models. Similar reduction can also be observed on the CPU of the Raspberry Pi devices.

For reference, the EMIT sensor collects up to 300 granules per day (and usually much less), and the NASA's planned SBG mission \cite{cawse2021nasa_SBG} will produce an estimate of about 10x more. These granules have 1280x1242 px and can be tiled into 100 tiles of 128x128 which we used as reference when measuring model runtimes.
With the estimated 300 granules and using the measurements from the Unibap machines, the mag1c matched filter baseline would run for almost 17 hours (with 203 seconds for each capture), while the proposed ConvUp variant of our HyperSegFormer would take just 2.5 hours (with 30 seconds for each capture). 
This time makes it more feasible to catch up with the estimated rate of data capture, we also assume that data loading could be handled on another thread in a delayed pipeline as in \cite{ruuvzivcka2018clusters}. Further reduction to just estimated 3.2 minutes (with 0.64 seconds for each capture) could be achieved when using the GPU on the Jetson Xavier board. This outlines the potential speed-up future space missions could leverage if using high-power compute units (even if these are turned on just for short periods of time). 

Finally, we note that while the reported results measure inference times of the methane leak detection models, these findings also generalise to our mineral identification models. Slower inference times can be expected due to the increased spectral dimensionality of the used data (285 bands for the mineral identification models in contrast to 86 bands used by the models for methane detection).

\section{Conclusion}
\label{sec:conclusion}

To summarise, our results indicate several interesting findings. 
Firstly, we release HyperspectralViTs, the Transformer based architectures adapted for the domain of hyperspectral data.
The adapted models outperform the default versions and baselines both in accuracy scores and in inference speeds on-board of devices that serve as a proxy for the compute environment of the ION-SCV004 satellite.

Secondly, we demonstrate new state-of-the-art results in methane detection using our end-to-end machine learning architectures adapted for hyperspectral data. Namely, on a newly created dataset of synthetic events in data from the EMIT sensor, we improve the F1 score by more than 27\% against the prior models depending on classical methane enhancement products from \cite{STARCOP} and by over 97\% in F1 score over classical matched filter baselines. 
When comparing with off-the-shelf performance, our proposed Hyperspectral variant of the SegFormer model improves the F1 score by over 23\% on this synthetic dataset.
We confirm our findings also on a benchmark dataset released by \cite{STARCOP}, where we improve the overall F1 score by almost 13\% and the AUPRC score by more than 17\%. Finally, on a newly created dataset of real methane leak events recorded by the EMIT sensor, we demonstrate, that it is possible to leverage models pretrained on synthetic data by fine-tuning them on smaller real datasets. Using the best performing model variant, we show that the finetuned model improves the F1 score by 6.9\% over models trained from scratch.

On a newly created dataset for the task of mineral identification, we note that our HyperSegFormer (ConvUpStride variant) improves the F1 score by over 3.5\% in contrast to the vanilla model. More notably, we show the influence of the retained output resolution. While the default versions of the Transformer based models lose details in predicted scenes, our adaptations greatly improve the quality of the predictions. We note that our results are conducted on an initial dataset with pseudo-ground truth labels made by the Tetracorder method \cite{clark2003imaging_Tetracorder}. However, with a manually verified dataset, it will be easy to retrain our models.
Our initial results pave the path for future experiments - to support this, we release our dataset with labels for all separate constituents.

Furthermore, our models could be leveraged as Foundation models initially trained on the detection of individual chemical components and later finetuned or adapted for a variety of downstream tasks (methane leak detection included). Gained performance on the general detection task could also help the other tasks of interest, as was exemplified in the case of multi-task learning \cite{kokkinos2017ubernet}.

In terms of inference speed, we achieve a 85.19\% faster performing model against the previously reported models and baselines of \cite{STARCOP}. We report the inference times measured on hardware which is a realistic proxy to the currently flying ION-SCV 004 satellite.
We highlight that with our model, one scene can be processed in around half a minute (6.7x times speedup against the baseline).
We also measure our models on several commercial off-the-shelf (COTS) devices, these exhibit further potential speedups and have started to be explored for space missions in \cite{swope2022benchmarking_ISS_snapdragon}. Namely, we show over 317x times speedup when using GPUs on the NVIDIA Jetson AGX Xavier board.

As a potential limitation of the proposed end-to-end \mbox{HyperspectralViT} models, they operate directly on \fix{the bands of} hyperspectral data and as such, they can't be used for zero-shot generalisation as for example the model of \cite{STARCOP}. 
We however note, that a similar simulated dataset can be created from representative data for any other satellite and later used to train end-to-end models.
Secondly, our model doesn't support additional tasks such as plume quantification or source point estimation. We however note, that the prediction of our model can be used as a binary mask to clean the noisy methane enhancement products, after which classical pipelines for these tasks can be used \cite{thorpe2016mapping_estimation}. 

\fixM{As a further limitation, we also note, that we used the L1B level of data processing in our experiments instead of the truly ``raw'' versions of the data (sometimes called L0). This is due to the fact that this data is not available for these sensors, in fact this is a common problem in the wider research of methods for on-board data processing. Only recently, small datasets for multispectral sensors were made available in the works of \cite{meoni2024_thraws}. However, works such as \cite{henriksen2019real} show that real-time correction and preprocessing of the L0 hyperspectral data is feasible on low-cost and low-compute platforms. Similarly, the work of \cite{castano2006onboardSVM} uses partially processed data (so-called level 0.5) on-board of the EO-1 spacecraft. Furthermore, works such as \cite{aybar2024onboard} demonstrate that tiny machine learning models can serve as simulators for fast translation between processing levels (in that case L1C to L2A of Sentinel-2 data). For real world deployment, we'd suggest two approaches: either using even partial data processing pipelines on-board, or training deep models with simulated L0-like data.}{R1Q1}

For future research, we'd like to explore further steps to make the models more efficient. Our proposed changes to the Transformer based models model can be combined with pruning reported in \cite{bai2022dynamically_pruneSegformer, yang2023pruning_segformers}. 
Specialised accelerators as those suggested by \cite{yousfia2024spikingStarcop} could also be leveraged.
Secondly, we'd also like to explore extension of our models with methods for uncertainty estimation. Methods such as model ensembles \cite{beluch2018power_ensembles}, or Monte Carlo Dropout \cite{gal2016dropout_mcdo} could be easily added to the proposed semantic segmentation models, as was previously demonstrated in \cite{ruuvzivcka2020deep}.
Thirdly, we note that the existing archives of EMIT data (currently with more than \fix{121k} captures totaling more than \fix{244} TB of data) offer unique opportunities to train large general foundation models on hyperspectral data, where one of the example tested tasks could also be methane detection. Models and datasets proposed in this paper could likely be leveraged for general learning tasks (in the direction outlined by our mineral identification task), but also for other downstream applications highlighted by the future planned missions such as NASA's SBG \cite{cawse2021nasa_SBG} or ESA's CHIME \cite{rast2021copernicus_CHIME}. 

Finally, we are releasing the three newly created datasets at \url{https://huggingface.co/previtus} split into several repositories - \href{https://huggingface.co/datasets/previtus/OxHyperRealCH4}{OxHyperRealCH4}, \href{https://huggingface.co/datasets/previtus/OxHyperSyntheticCH4}{OxHyperSyntheticCH4} and \href{https://huggingface.co/datasets/previtus/OxHyperMinerals_Train}{OxHyperMinerals}. The code and the pre-trained models are on Github at \url{https://github.com/previtus/HyperspectralViTs}.

\section*{Acknowledgments}
We would like to thank D-Orbit and Unibap for access to the SpaceCloud® Hardware when measuring our models inference speeds on realistic hardware proxy of a real satellite environment.

\bibliographystyle{IEEEtran}
\bibliography{IEEEabrv,bib}
%

% \newpage

\section{Biography Section}
 
\vspace{11pt}

\vspace{-33pt}
\begin{IEEEbiography}[{\includegraphics[width=1in,height=1.25in,clip,keepaspectratio]{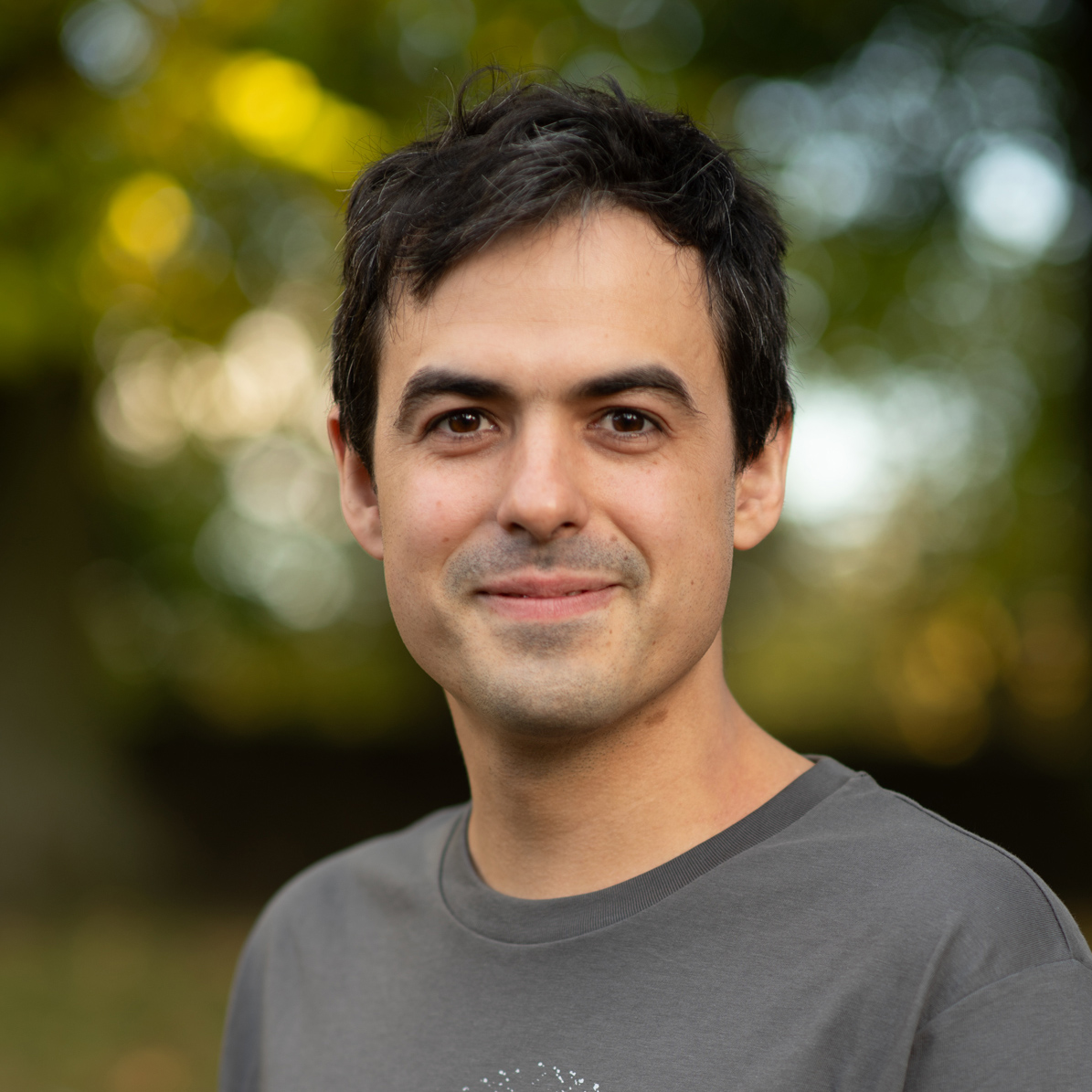}}]{Vít Růžička} received the B.Sc. and M.Sc. degrees in Computer Science from the Czech Technical University, Prague, Czechia in 2014 and 2017. In April 2025, he received his Ph.D. degree in Computer Science at the University of Oxford, Oxford, U.K.

He was a Visiting Researcher with the European Space Agency (2023) and a Consultant for the United Nations Environment Programme (2024). His research interests include machine learning applied to remote sensing applications and tasks connected to climate change and on-board deployment on low compute systems, such as satellites.
\end{IEEEbiography}

\vspace{-33pt}

\begin{IEEEbiography}[{\includegraphics[width=1in,height=1.25in,clip,keepaspectratio]{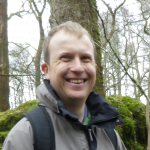}}]{Andrew Markham} is a professor of Computer Science at the University of Oxford, UK. He received his BSc (Hons) (2004) and PhD (2008) both from the University of Cape Town, South Africa in Electrical Engineering. 

He has published over 200 papers and works in the broad area of cyberphysical system. His research interests are on the application of sensing, systems, and signal processing to increased sustainability, conservation, and autonomy.
\end{IEEEbiography}

\vspace{11pt}

\vfill

\end{document}